\documentclass[conference]{IEEEtran}
\IEEEoverridecommandlockouts
\usepackage{cite}
\usepackage{amsmath,amssymb,amsfonts}
\usepackage{algorithmic}
\usepackage{graphicx}
\usepackage{textcomp}
\usepackage{xcolor}
\def\BibTeX{{\rm B\kern-.05em{\sc i\kern-.025em b}\kern-.08em
    T\kern-.1667em\lower.7ex\hbox{E}\kern-.125emX}}
\begin{document}

\title{A Fast Location Algorithm for Very Sparse Point Clouds Based on Object Detection
}

\author{\IEEEauthorblockN{1\textsuperscript{st} Shiyu Fan}
\IEEEauthorblockA{\textit{dept. name of organization (of Aff.)} \\
\textit{name of organization (of Aff.)}\\
City, Country \\
email address or ORCID}
}

\maketitle

\begin{abstract}
Limited by the performance factor, it is arduous to recognize target object and locate it in Augmented Reality (AR) scenes on low-end mobile devices, especially which using monocular cameras. In this paper, we proposed an algorithm which can quickly locate the target object through image object detection in the circumstances of having very sparse feature points. We introduce YOLOv3-Tiny to our algorithm as the object detection module to filter the possible points and using Principal Component Analysis (PCA) to determine the location. We conduct the experiment in a manually designed scene by holding a smartphone and the results represent high positioning speed and accuracy of our method.
\end{abstract}

\begin{IEEEkeywords}
augment reality, object detection, YOLOv3-Tiny, PCA
\end{IEEEkeywords}

\section{Introduction}
Augmented Reality (AR) is a research hot spot on mobile smart devices gradually. More developers utilize AR to develop diverse mobile games and video effects. AR integrates computer three-dimensional reconstruction with computer vision and establishes the mapping relationship between the real world and the screen, so that the 3D models we want to draw can be displayed on the screen as if they are attached to real objects. In many circumstances, developer need detect one specific target’s position in real world and transform it to virtual rendering space. Owing to the rapid development of Simultaneous Localization and Mapping (SLAM) and computer vision, researchers apply a variety of different method to locate the target.

In most of SLAM system, the plane can be easily identified and tracked. In this case, developer can put object on the plane or hang something on the wall. Meanwhile, researchers develop other method to detect object without regular shape. AR marker is a 
However, no matter marker system or scanning system, most of current target location solutions rely on extra markup or preprocessing of the scene. These solutions cannot handle generalized or diverse objects moreover.

To solving the problem that mentioned before, we propose a new method which can be applied easily in mobile device. The process of this method combines object detection algorithm based on neural network with Principal Component Analysis (PCA) and it concludes three main parts. Firstly, the object recognition network outputs the bounding box of target from the camera stream. Then, the sparse feature points extracted by SLAM are projected to the screen space and the bounding box filter out the effective points. The last and most important step is applying PCA to calculate the approximate location and direction.

This method is test on the real scene experiment captured by mobile device and the result verifies the feasibility of this method. It indicates that our system can successfully recognize the target we set and locate the transform matrix of the target. 

\section{Related Work}

For target detection and positioning, the current AR system proposes different solutions according to its own characteristics. For the detection and positioning of flat objects, most of the algorithms in the industry are template matching for the feature points of flat objects. These methods extract the SIFT\cite{} feature points from target and calculate the transformation matrix through the deformation of target to realize the positioning function\cite{}. Another way of target recognition and positioning is through learning. The point cloud information generated by slam can be used by neural networks. In recent years, a large number of networks for target detection and location based on the point cloud generated by lidar have been proposed, e.g. PointNet\cite{}, MV3D\cite{} and Point RCNN\cite{}. However, all these methods rely on the LiDAR.

In this paper, We used an image-based approach for target positioning without any depth information. Object detection algorithms has two types: one is traditional machine learning type and the other is deep neural network. For those using neural network, they can be divided into one-stage and two-stage. Two-stage refers to the detection algorithm that needs to be completed in two steps. First, the proposal areas need to be cropped from the original image and then the network classify all these areas to select the best match. Algorithms grounded on this principle include R-CNN\cite{}, Faster R-CNN\cite{} and so on. However, due to the huge cost for classifying all regions, the hardware resources requirement for this type of algorithm are relatively high, even Faster R-CNN operate at \\ speed, which means it cannot be applied for the real-time operation. One-stage detectors cut down the step of generating proposal region and extract image features only once, SSD\cite{} and YOLO\cite{} are the most typical examples.YOLO divide one image into several grids, each grid is responsible for predicting the target whose center is inside this grid. This means that each grid predicts several bounding box's position and confidence at the same time. The first generation of YOLO has the defect of poor detection of small targets. In subsequent version, YOLO replace the backbone from VGG-16 to DarkNet-19 and improve the performance of speed and accuracy\cite{}. In YOLOv3\cite{}, the ideal of residual network is introduced and three feature maps of different scales are used for object detection.
\section{Methodology}
The main process of our method is shown as in Fig.1. We use ARkit in our pipeline to point clouds generation. Note that other monocular SLAM like ARCore can also be used in this stage.

\begin{figure}[htbp]
\centerline{\includegraphics[width=0.48\textwidth]{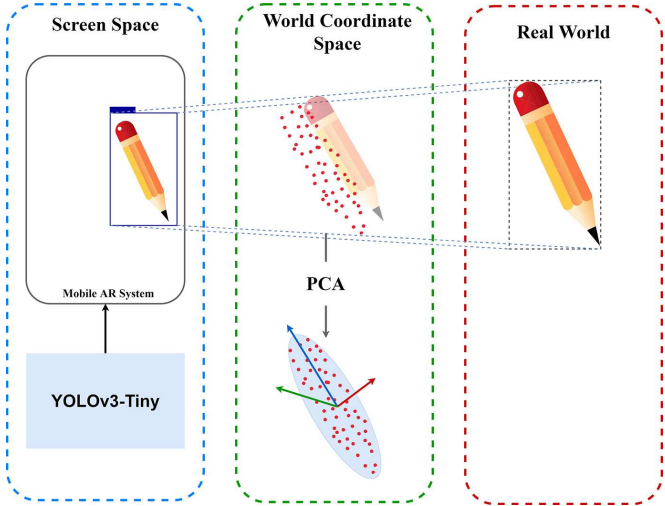}}
\caption{Example of a figure caption.}
\label{fig}
\end{figure}

The object detection module-YOLOv3-Tiny is modified by YOLOv3, which is a quiet large network and not suitable for the mobile platform. In this case YOLOv3-Tiny has implement the following changes. First,the backbone of YOLOv3-Tiny is simplified. The original backbone network implemented in YOLOv3 is DarkNet-53, which has 53 convolution layers. In YOLOv3-Tiny, the structure of backbone is a 7-layers of convolution and max-pooling. Secondly, comparing to YOLOv3, YOLOv3-Tiny only retains two independence prediction branches with 16x16 and 32x32 resolutions. Furthermore, YOLOv3-Tiny apply quantization strategy to reduce the BFLOPS. In this network, the weights are stored in low precision as INT8 type. Comparing to the YOLOv3-SPP3, YOLOv3-Tiny successfully decrease parameter's amount from 63.9M to 8.7M. When with the same input size as 412, the BFLOPS of YOLOv3-Tiny is 7.69\% of YOLOv3 has.

\begin{table}[htbp]
\caption{Table Type Styles}
\begin{center}
\begin{tabular}{ccccc}
\hline
\textit{\textbf{\begin{tabular}[c]{@{}c@{}}Model\\ Name\end{tabular}}} & \textit{\textbf{\begin{tabular}[c]{@{}c@{}}Input\\ Size\end{tabular}}} & \textit{\textbf{\begin{tabular}[c]{@{}c@{}}Parameter\\ Size\end{tabular}}} & \textit{\textbf{\begin{tabular}[c]{@{}c@{}}Model\\ Size\end{tabular}}} & \textit{\textbf{FPS}} \\ \hline
YOLOv3                                                                 & 416x416                                                                & 65.3M                                                                      & 263MB                                                                  & 5.32                  \\
YOLOv3-Tiny INT8                                                       & 416x416                                                                & 8.7MB                                                                      & 8.9MB                                                                  & 222.59                \\ \hline
\end{tabular}
\label{tab1}
\end{center}
\end{table}

Trained YoloV3-Tiny network return the target box in image coordinate system. The coordinate of this box can be easily transformed from image coordinate system to a  normalized device coordinate(NDC) system. The 

Meanwhile, the world coordinates of the sparse point cloud can be converted into the screen coordinate system. With the following steps, we can filter the possible point cloud data related to the target object. The transformation of an object from world coordinates to NDC space needs to undergo camera view matrix $M_{view}$ transformation, orthogonal projection matrix $M_{Otho,norm}$ transformation and perspective projection $M_{proj}$ matrix transformation. Assuming $P$ is one point of the cloud which has the world coordinate $P_{world}(x_{w},y_{w},z_{w})$ $M_{p}$. Generally, point $P$'s position in NDC space can be obtained by the following formula:

\begin{equation}
P_{ndc}=M_{proj}M_{otho,norm}M_{view}^{-1}P_{w}
\end{equation}

After calculating all the raw points position of NDC coordinate, the points inside the target box will be analyzed in the next step. 

Following the previous steps, we get the point cloud set $W\left \{ P_{1},P_{2},P_{3},...,P_{n} \right \}$ located in the target box. The center $C$ of target object can be obtained by the mean of all points‘ coordinate in $W$ set.

\begin{equation}
(x_{c},y_{c},z_{c})= (\frac{\sum_{1}^{n}x_{n}}{n},\frac{\sum_{1}^{n}y_{n}}{n},\frac{\sum_{1}^{n}z_{n}}{n})
\end{equation}
With the center of the point cloud data, we can use PCA to calculate the principal component vector of the point cloud data, and to roughly calculate the direction vector of the point cloud. To ensure that the first principal component is the direction of maximum variance, data matrix $X_{w}\in\mathbb{R}^{3\times n}$.
\begin{equation}
X_{w}= \begin{bmatrix}
x_{1}-x_{c} & x_{2}-x_{c} & ...& x_{n}-x_{c}\\ 
y_{1}-y_{c} & y_{2}-y_{c} & ...& y_{n}-y_{c}\\
z_{1}-z_{c} & z_{2}-z_{c} & ...& z_{n}-z_{c}\\ 
\end{bmatrix}
\end{equation}
By calculating the covariance matrix, and then according to the three eigenvectors of the covariance matrix, the deflection angle of the point cloud data can be determined.
\begin{equation}
\Sigma_{w}\in\mathbb{R}^{3\times 3}=\frac{1}{n-1}X_{w}X_{w}^{T}
\end{equation}
\begin{equation}
\left |\Sigma_{w}-\lambda I   \right |=0
\end{equation}
The three $\lambda_{1}$, $\lambda_{2}$ and $\lambda_{3}$ of matrix $\Sigma_{w}$ obtained by the solution correspond to the three eigenvectors $\lambda_{1}$, $\lambda_{2}$ and $\lambda_{3}$.

The above operations can also be replaced by singular value decomposition(SVD):
\begin{equation}
X_{w}=U\Sigma V^T    
\end{equation}
where $U$, $\Sigma$ are , and $V$ is the combination of $\lambda_{1}$, $\lambda_{2}$ and $\lambda_{3}$.

In the process of using the above algorithm, the following issues need to be pay attention. First, due to the RGB cameras used by most mobile devices, point clouds require dozens of frames to be created. This means that there must be enough target feature points to be generated before starting the process. In this paper, we set a threshold $N$ of feature points' number for starting detection process. Secondly, since the eigenvectorsr can take two different values, we need to set the value range of the vector in advance.

\section{Implementation}
With our system integrated on IOS Scenekit platform, 
\section{Experiment Result}

\section{Limitations}
Due to the limitations of the algorithm, our method only works well on regular objects with certain directivity. Otherwise, the scale of YOLOv3-Tiny is too large for some mobile devices that it cannot perform real-time object detection.

\section{Conclusion}
On most mobile platforms, the point cloud generated by the AR system is extremely sparse and it is completely impossible to obtain effective target information from these point cloud data. The method we propose in this paper can successfully recognize the target on mobile devices and acquire the position of the target in the rendering world space coordinate system and the possible rotation angle.

\section*{References}
Continue...

\end{document}